# AI Discovering a Coordinate System of Chemical Elements: Dual Representation by Variational Autoencoders

**Alex Glushkovsky**
alexglu@gmail.com

**Abstract**

The periodic table is a fundamental representation of chemical elements that plays essential theoretical and practical roles. The research article discusses the experiences of unsupervised training of neural networks to represent elements on the 2D latent space based on their electron configurations. To emphasize chemical properties of the elements, the original data of electron configurations has been realigned towards valence orbitals. Recognizing seven shells and four subshells, the input data has been arranged as 7x4 images. Latent space representation has been performed using a convolutional beta variational autoencoder (beta-VAE). Despite discrete and sparse input data, the beta-VAE disentangles elements of different periods, blocks, groups, and types. The unsupervised representation of elements on the latent space reveals pairwise symmetries of periods and elements related to the invariance of quantum numbers of corresponding elements. In addition, it isolates outliers that turned out to be known cases of Madelung's rule violations for lanthanide and actinide elements. Considering the generative capabilities of beta-VAE, the supervised machine learning has been set to find out if there are insightful patterns distinguishing electron configurations between real elements and decoded artificial ones. Also, the article addresses the capability of dual representation by autoencoders. Conventionally, autoencoders represent observations of input data on the latent space. By transposing and duplicating original input data, it is possible to represent variables on the latent space which can lead to the discovery of meaningful patterns among input variables. Applying that unsupervised learning for transposed data of electron configurations, the order of input variables that has been arranged by the encoder on the latent space has turned out to exactly match the sequence of Madelung's rule.

## 1 Introduction

In 1869, Mendeleev published his discovery of organizing chemical elements as a periodic table with distinct order based on periodic and grouping properties. This discovery is a phenomenal example of human intelligence to arrange scattered elements into organized representation (Figure 1). It has had remarkable theoretical and practical values for over 150 years and counting. Incredibly, he predicted the existence of elements and their properties by noting missing cells in his table and then later witnessed some of them being discovered on our planet.

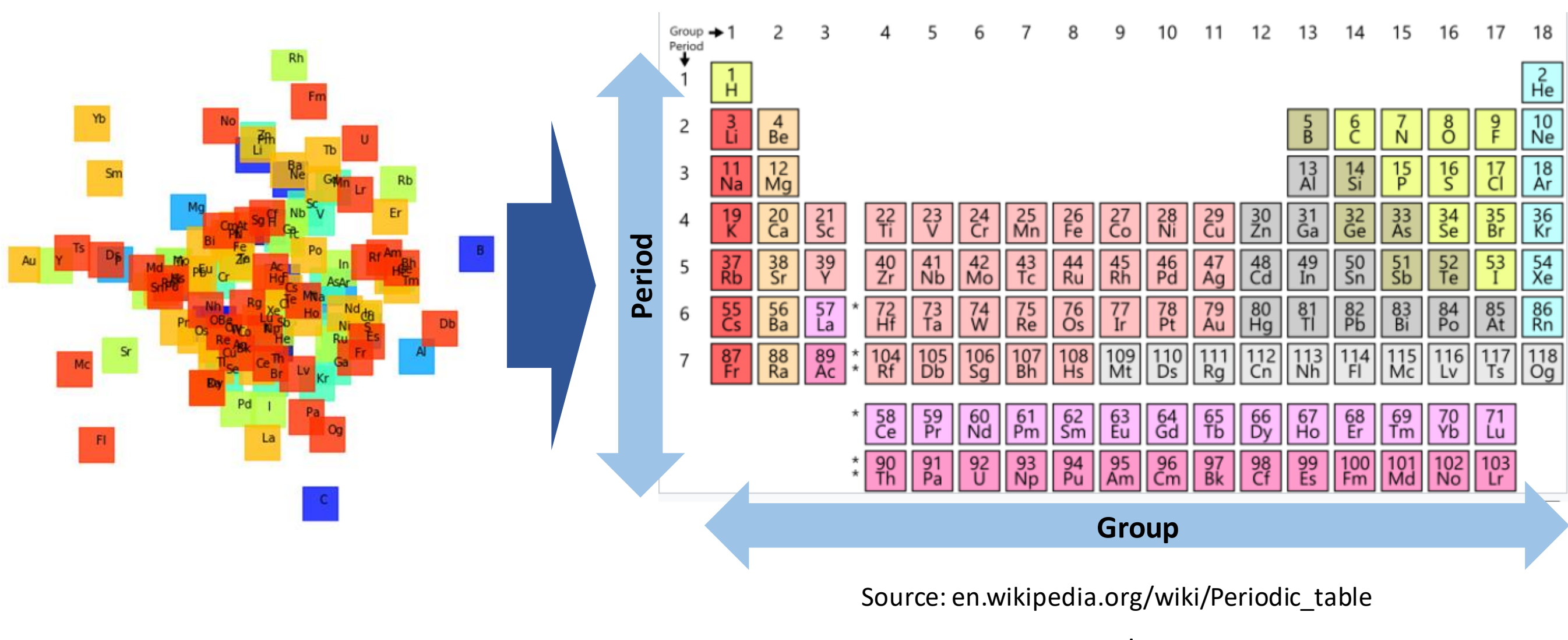

Figure 1. Scattered presentation of chemical elements (a) and an organized periodic table example (b)

There are numerous versions of the periodic table having different sizes such as short or long column tables; different shapes such as spiral or zigzag; different dimensions such as 2D or 3D – just to name a few. Seven hundred different versions of the periodic table published within 100 years after Mendeleev's original publication were compiled by (Mazurs, 1974).

Such attention to the representations of chemical elements reflects their exceptional importance. Elements are instrumental in understanding fundamentals of our universe. It is in focus by different sciences: physics, including quantum physics, chemistry, and even mathematics – for instance, prime numbers are deep-rooted in the electron configurations of the elements.

Applications of AI and underlying machine learning algorithms in chemistry have become a very intensive area of research and development, especially deep generative modeling for molecular design. For example, the review that addresses applications of generative modeling algorithms in designing and optimizing molecules has been presented in (Elton *et al*, 2019) and comprises 45 papers published in just the last two years. Deep learning approaches, such as ElemNet, are able to predict unknown compounds (Jha *et al*, 2018). Special interest is given to molecular latent space discovery with variational autoencoders, such as those described in (Shervani-Tabar and Zabaras, 2020).

Unprecedented recent developments of machine learning algorithms and their already successful applications in chemistry triggers the question: "Are unsupervised algorithms capable of meaningfully organizing chemical elements?" I.e., "Can they "rediscover" the periodic table that the human brain was able to envision?" Or, even more ambitious: "Can AI unpack some principles that bind electron configurations of the elements?".

Thus, the application of an unsupervised Self-Organizing Map (SOM) has been applied in order to classify elements based on properties known at Mendeleev's time (Lemes and Pino, 2011). Unsupervised Atom2Vec learns basic properties of atoms based on atom-environment pairs from the database of compounds and materials. It was able to rediscover the periodic table (Zhou *et al*, 2018). The Smooth Overlap of Atomic Positions (SOAP) machine learning algorithm, that is widely used in chemistry, reduces the dimensionality of the representation of elements and shows it is similar to the conventional periodic table (Willatt *et al*, 2019). To recreate the periodic table, the unsupervised Periodic Table Generator (PTG) has been developed based on generative topographic mapping (Kusaba *et al*, 2019). To reduce missing values, training has been done based on the first 54 elements and 39 features describing properties, such as melting point temperatures or electronegativity. The 2D array and 3D conical table with learned periodicity have been created by applying PTG.

The periodic table of elements can be seen as a representation that is arranged by atomic numbers and that disentangles periods and groups.

Recently, numerous great approaches have been developed addressing representation on the latent space. It includes embedding, generative variational autoencoders, and disentanglement, such as t-SNE, β-VAE, InfoVAE, InfoGAN (Van der Maaten and Geoffrey, 2008; Bengio, 2013; Kingma and Welling, 2014; Chen *et al*, 2016; Makhzani *et al*, 2016; Higgins *et al*, 2017; Zhao *et al*, 2018). Special attention is given to representations of discrete cases, such as (Bengio *et al*, 2013; Jang *et a*l, 2017; Rolf, 2017; Dupont, 2018).

In this paper, generative β-VAE has been applied considering the regularization term of Kullback–Leibler divergence from standard isotropic multivariate normal distribution. The unsupervised generative β-VAE has the following features: it compresses information of inputs into a latent vector, disentangles underlying characteristics of the input objects, and generates a decoded output for a given location on the latent space. In addition, the beta parameter, that is a Lagrangian multiplier, provides flexibility to balance between decoded accuracy and disentanglement (Higgins *et al*, 2017).

The paper discusses experiences of unsupervised training of β-VAE to represent chemical elements based on their electron configurations on the 2D latent space forcing disentanglement. In addition, it addresses the capability of dual representation by autoencoders that underlines Madelung's rule.

## 2 Data and Challenges

The experiments of unsupervised rediscovery of the periodic system of elements have been performed based on electron configurations (Shultis and Faw, 2007).

The first trail has been performed using data concerning the number of electrons per each of the seven shells for 118 known elements as it is shown in Table 1, a. The complete data sources of electron configurations and chemical elements data used in this paper are from (https://en.wikipedia.org/wiki/Electron_configurations_of_the_elements_(data_page)) or corresponding Wikipedia sites.

The described input data has the number of challenges considering the usage of machine learning algorithms: discrete and sparse data, limited integer values of data, and small number of observations – only 118 elements. To overcome these issues a simple approach of duplicating input data 100 times has been applied with and without adding Gaussian normally distributed random noise ~ $\alpha \cdot N(0,1)$. Noting that the noise violates quantum physics' principles, it has been introduced in order to improve robustness of the modeling process and to prevent overfitting (Guozhong An, 1996).

It turned out that β-VAE could be run against that duplicated dataset and that adding Gaussian noise has no noticeable effects on the obtained results.

| Element | | Shell | | | | | | |
|---|---|---|---|---|---|---|---|---|
| Num | Name | 1 | 2 | 3 | 4 | 5 | 6 | 7 |
| 1 | H | 1 | | | | | | |
| 2 | He | 2 | | | | | | |
| 3 | Li | 2 | 1 | | | | | |
| 4 | Be | 2 | 2 | | | | | |
| 5 | B | 2 | 3 | | | | | |
| 6 | C | 2 | 4 | | | | | |
| 7 | N | 2 | 5 | | | | | |
| 8 | O | 2 | 6 | | | | | |
| 9 | F | 2 | 7 | | | | | |
| 10 | Ne | 2 | 8 | | | | | |
| 11 | Na | 2 | 8 | 1 | | | | |
| 12 | Mg | 2 | 8 | 2 | | | | |
| 13 | Al | 2 | 8 | 3 | | | | |
| 77 | Ir | 2 | 8 | 18 | 32 | 15 | 2 | |
| 116 | Lv | 2 | 8 | 18 | 32 | 32 | 18 | 6 |
| 117 | Ts | 2 | 8 | 18 | 32 | 32 | 18 | 7 |
| 118 | Og | 2 | 8 | 18 | 32 | 32 | 18 | 8 |

a

| Element | | _Shell | | | | | | |
|---|---|---|---|---|---|---|---|---|
| Num | Name | _7 | _6 | _5 | _4 | _3 | _2 | _1 |
| 1 | H | | | | | | | 1 |
| 2 | He | | | | | | | 2 |
| 3 | Li | | | | | | 2 | 1 |
| 4 | Be | | | | | | 2 | 2 |
| 5 | B | | | | | | 2 | 3 |
| 6 | C | | | | | | 2 | 4 |
| 7 | N | | | | | | 2 | 5 |
| 8 | O | | | | | | 2 | 6 |
| 9 | F | | | | | | 2 | 7 |
| 10 | Ne | | | | | | 2 | 8 |
| 11 | Na | | | | | 2 | 8 | 1 |
| 12 | Mg | | | | | 2 | 8 | 2 |
| 13 | Al | | | | | 2 | 8 | 3 |
| 77 | Ir | | 2 | 8 | 18 | 32 | 15 | 2 |
| 116 | Lv | 2 | 8 | 18 | 32 | 32 | 18 | 6 |
| 117 | Ts | 2 | 8 | 18 | 32 | 32 | 18 | 7 |
| 118 | Og | 2 | 8 | 18 | 32 | 32 | 18 | 8 |

b

Table 1. Original (a) and realigned (b) data of electron configurations of the elements across the seven shells

An example of the chemical elements representation on the latent space based on the original electron configurations along the seven shells (Table 1, a) is shown in Figure 2. It can be noted that the representation has low information value other than a sequential order of elements based on their atomic numbers. Thus, the latent space manifold has no separations along periods (color coded) and has no significant irregularities except for some sharp corners.

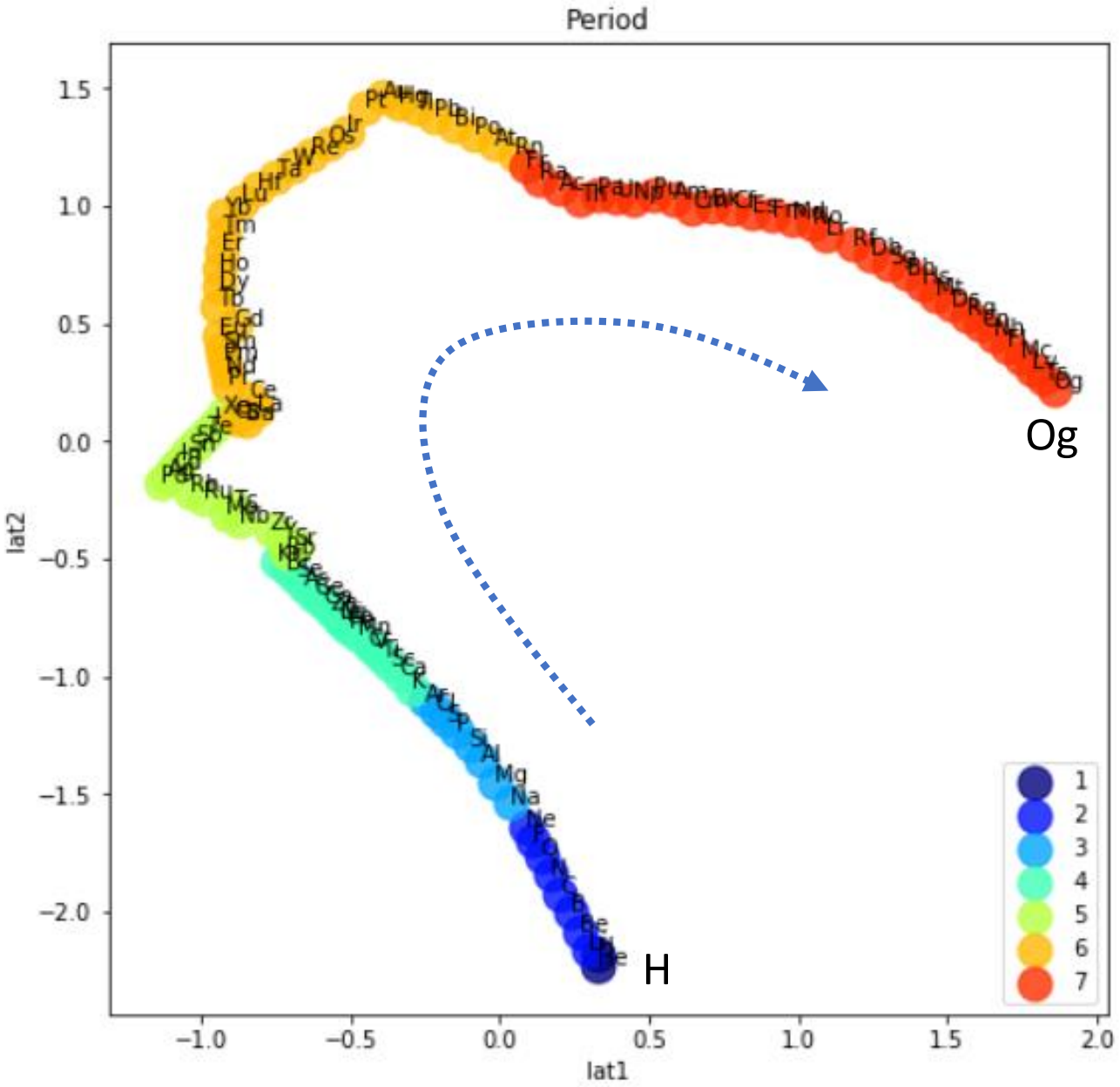

Figure 2. Example of the chemical elements representation on the latent space based on the original data of electron configurations

The representations on the latent space included 2D and 3D experiments. It turns out that 3D latent space provides a smaller loss when keeping other hyperparameters unchanged, but significantly challenges interpretations. The paper is limited to 2D representations but, as a research instrument, the 3D or even higher dimensionalities might be useful. Other experimentations have been performed concerning the architecture of neural networks and tuning of the beta and other hyperparameters. The finalized versions of the applied neural networks are provided in the Appendix.

# 3 Input Data Realignment as Feature Engineering

In order to emphasize chemical properties of elements and trigger disentanglement, a simple data transformation has been applied. The original data shown in Table 1 (a) is aligned towards a nuclear: the lower the number of the shell, the closer it is to the center of the atom. However, acknowledging the presence of valence electrons that are located on the outermost orbitals and their essential roles in chemical properties of the elements (Cracolice and Peters, 2016), the data has been realigned towards the outer orbitals as shown in Table 1 (b) (see Appendix, 1). The arrangement of elements according to their valences has been presented as a left-step series table by (Mazurs, 1974).

After applying that simple yet crucial data realignment, the representation of elements on the latent space shows very distinct patterns of disentangled periods and groups (Figure 3). Essentially, this result rediscovers periodic table properties and has been obtained by unsupervised machine learning algorithm. However, it should be noted, that the learned representation has been obtained based on modern knowledge of atom structure, specifically electron configurations (Table 1, a), that were unknown during Mendeleev's time.

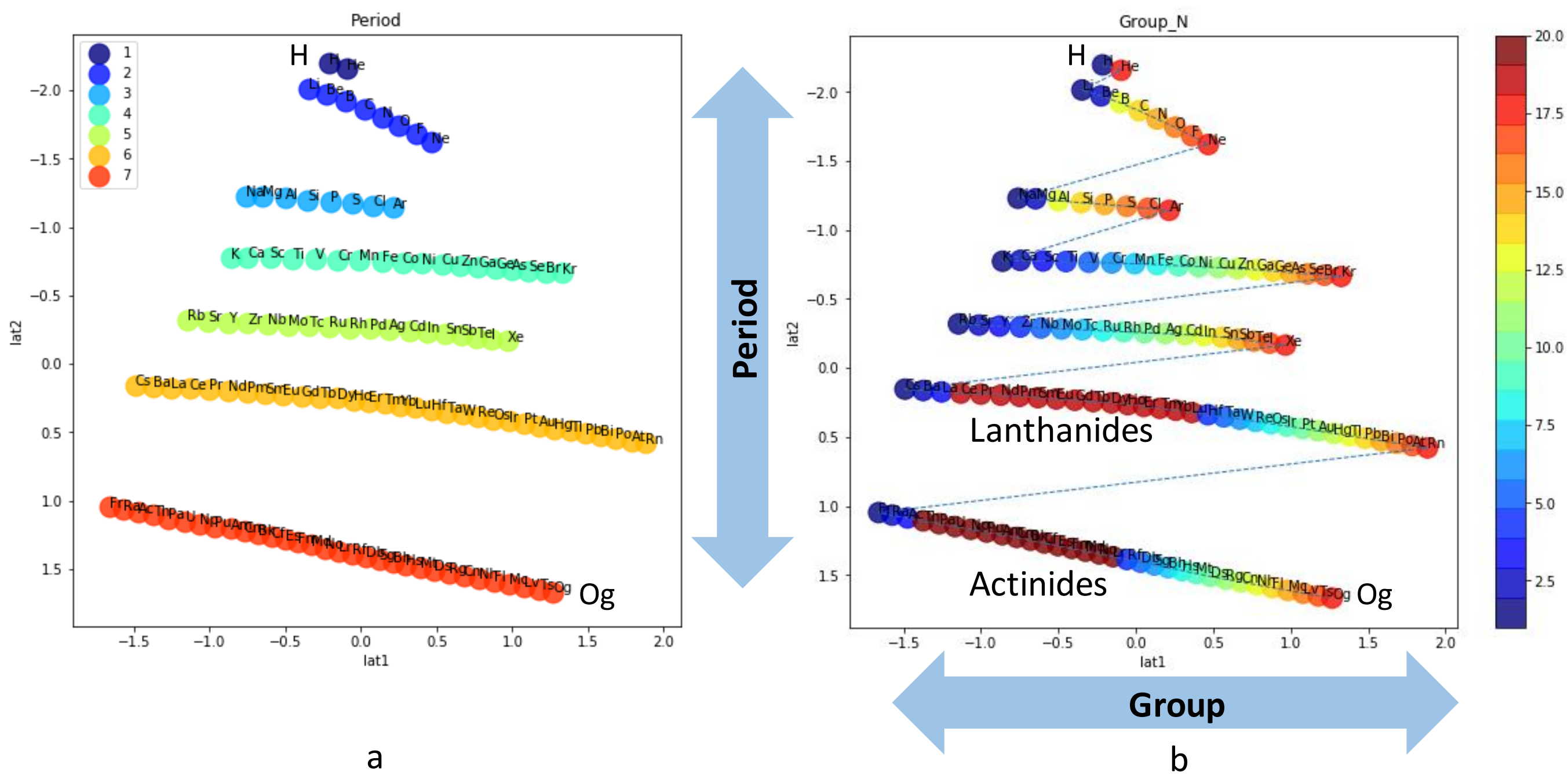

Figure 3. Example of the latent space representation based on realigned data towards valence orbitals (Table 1, b) with periods (a) and groups (b) disentanglements

It should be noted that machine learning is a stochastic process and there are variances of representations from run to run and, therefore, the shown figures of the latent space manifolds are just examples of some training outcomes.

# 4 Input Data with Subshells Levels

Encouraged by the obtained results, the next experiment has been performed by adding data of electron configurations on the subshell levels "s", "p", "d", and "f". Original table is shown in Table 2 (a) and it has 19 input variables. Similar to the above approach, that data has been realigned towards valence orbitals as shown in Table 2 (b) (see Appendix, 1).

The realignment shifts the shells of the outermost orbitals to be grouped together. A new artificial shell number has been assigned starting from the outermost orbital with increasing numbers towards nuclear, while not changing the subshell identifications. This realignment process has been repeated for the next shell to the most outer orbitals, and so on. In addition, this realignment incorporates the (n-1) for "d" and (n-2) for "f" orbitals shift rules affecting transition metals, lanthanides, and actinides (Charles and Madelung, 1936).

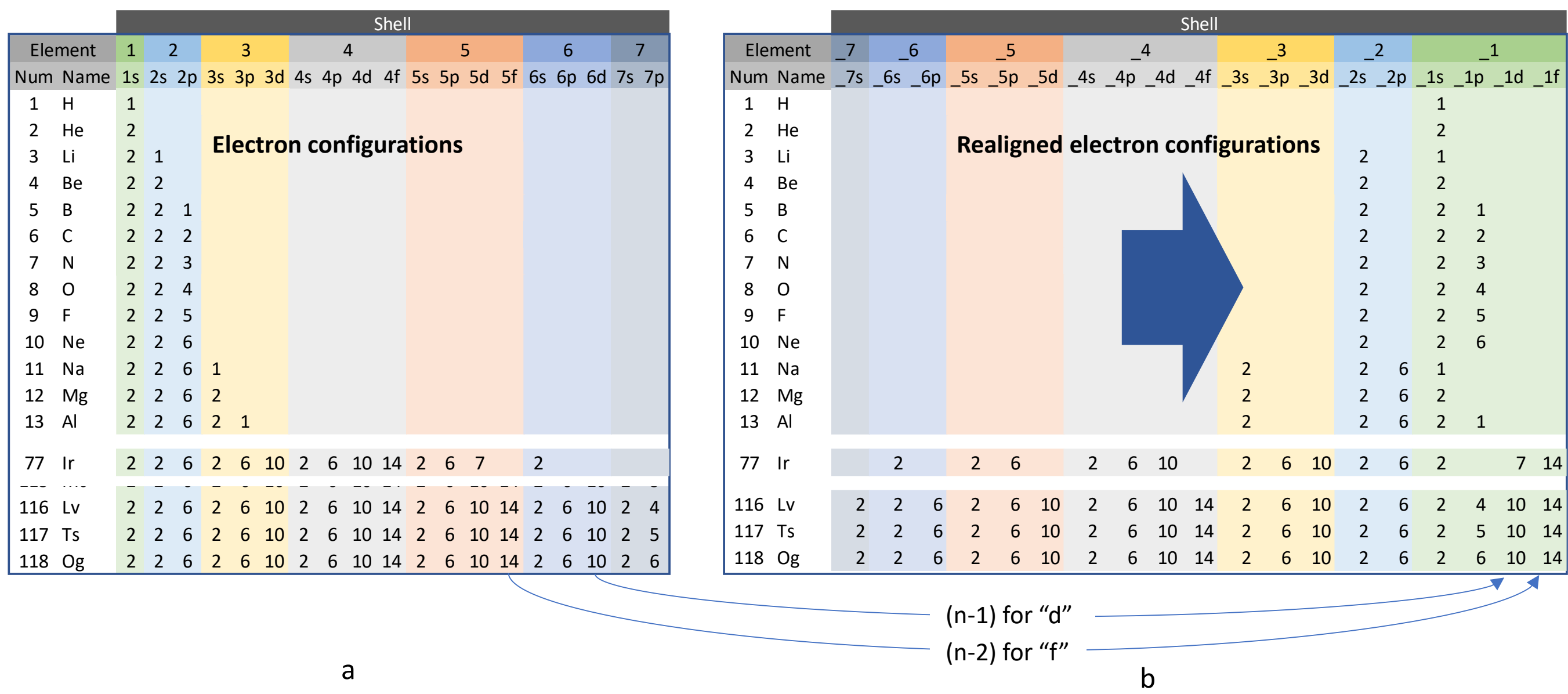

| | | Shell | | | | | | | | | | | | | | | | | | |
|---|---|---|---|---|---|---|---|---|---|---|---|---|---|---|---|---|---|---|---|---|
| Element | | 1 | 2 | | 3 | | | 4 | | | | 5 | | | | 6 | | | 7 | |
| Num | Name | 1s | 2s | 2p | 3s | 3p | 3d | 4s | 4p | 4d | 4f | 5s | 5p | 5d | 5f | 6s | 6p | 6d | 7s | 7p |
| 1 | H | 1 | | | | | | | | | | | | | | | | | | |
| 2 | He | 2 | | | | | | | | | | | | | | | | | | |
| 3 | Li | 2 | 1 | | | | | | | | | | | | | | | | | |
| 4 | Be | 2 | 2 | | | | | | | | | | | | | | | | | |
| 5 | B | 2 | 2 | 1 | | | | | | | | | | | | | | | | |
| 6 | C | 2 | 2 | 2 | | | | | | | | | | | | | | | | |
| 7 | N | 2 | 2 | 3 | | | | | | | | | | | | | | | | |
| 8 | O | 2 | 2 | 4 | | | | | | | | | | | | | | | | |
| 9 | F | 2 | 2 | 5 | | | | | | | | | | | | | | | | |
| 10 | Ne | 2 | 2 | 6 | | | | | | | | | | | | | | | | |
| 11 | Na | 2 | 2 | 6 | 1 | | | | | | | | | | | | | | | |
| 12 | Mg | 2 | 2 | 6 | 2 | | | | | | | | | | | | | | | |
| 13 | Al | 2 | 2 | 6 | 2 | 1 | | | | | | | | | | | | | | |
| 77 | Ir | 2 | 2 | 6 | 2 | 6 | 10 | 2 | 6 | 10 | 14 | 2 | 6 | 7 | | 2 | | | | |
| 116 | Lv | 2 | 2 | 6 | 2 | 6 | 10 | 2 | 6 | 10 | 14 | 2 | 6 | 10 | 14 | 2 | 6 | 10 | 2 | 4 |
| 117 | Ts | 2 | 2 | 6 | 2 | 6 | 10 | 2 | 6 | 10 | 14 | 2 | 6 | 10 | 14 | 2 | 6 | 10 | 2 | 5 |
| 118 | Og | 2 | 2 | 6 | 2 | 6 | 10 | 2 | 6 | 10 | 14 | 2 | 6 | 10 | 14 | 2 | 6 | 10 | 2 | 6 |

| | | Shell | | | | | | | | | | | | | | | | | | |
|---|---|---|---|---|---|---|---|---|---|---|---|---|---|---|---|---|---|---|---|---|
| Element | | _7 | _6 | | _5 | | | _4 | | | | _3 | | | _2 | | _1 | | | |
| Num | Name | _7s | _6s | _6p | _5s | _5p | _5d | _4s | _4p | _4d | _4f | _3s | _3p | _3d | _2s | _2p | _1s | _1p | _1d | _1f |
| 1 | H | | | | | | | | | | | | | | | | 1 | | | |
| 2 | He | | | | | | | | | | | | | | | | 2 | | | |
| 3 | Li | | | | | | | | | | | | | | 2 | | 1 | | | |
| 4 | Be | | | | | | | | | | | | | | 2 | | 2 | | | |
| 5 | B | | | | | | | | | | | | | | 2 | | 2 | 1 | | |
| 6 | C | | | | | | | | | | | | | | 2 | | 2 | 2 | | |
| 7 | N | | | | | | | | | | | | | | 2 | | 2 | 3 | | |
| 8 | O | | | | | | | | | | | | | | 2 | | 2 | 4 | | |
| 9 | F | | | | | | | | | | | | | | 2 | | 2 | 5 | | |
| 10 | Ne | | | | | | | | | | | | | | 2 | | 2 | 6 | | |
| 11 | Na | | | | | | | | | | | 2 | | | 2 | 6 | 1 | | | |
| 12 | Mg | | | | | | | | | | | 2 | | | 2 | 6 | 2 | | | |
| 13 | Al | | | | | | | | | | | 2 | | | 2 | 6 | 2 | 1 | | |
| 77 | Ir | | 2 | | 2 | 6 | | 2 | 6 | 10 | | 2 | 6 | 10 | 2 | 6 | 2 | | 7 | 14 |
| 116 | Lv | 2 | 2 | 6 | 2 | 6 | 10 | 2 | 6 | 10 | 14 | 2 | 6 | 10 | 2 | 6 | 2 | 4 | 10 | 14 |
| 117 | Ts | 2 | 2 | 6 | 2 | 6 | 10 | 2 | 6 | 10 | 14 | 2 | 6 | 10 | 2 | 6 | 2 | 5 | 10 | 14 |
| 118 | Og | 2 | 2 | 6 | 2 | 6 | 10 | 2 | 6 | 10 | 14 | 2 | 6 | 10 | 2 | 6 | 2 | 6 | 10 | 14 |

Table 2. Electron configurations across shells and subshells per elements: (a) original data and (b) realigned towards valence orbitals

Essentially, this realignment produces the correct electron configuration according to Madelung's rule (Madelung, 1936). For example, the 77th element Iridium has the following electron configuration based on Madelung's rule:

$$1s^2 2s^2 2p^6 3s^2 3p^6 3d^{10} 4s^2 4p^6 4d^{10} 5s^2 5p^6 4f^{14} 5d^7 6s^2$$

That configuration is correctly reflected in the transformed input data (Table 2, b).

In addition, the outermost shell of that element (_1: $4f^{14}5d^7 6s^2$) is aligned with the outermost shell of other elements, for example, the 12th element Magnesium: (_1: $3s^2$). The prefix "_" means that its' realigned configuration is towards the outermost valence orbitals.

This realignment process is feature engineering that has two essential attributes: (1) input features are regrouped emphasizing chemical properties of elements, and (2) the result of such realignment is uniquely reproduceable.

The data presented in Table 2, b can be viewed as having two dimensions: the shell number (1:7) and four subshell identifiers: "s", "p", "d", and "f". That notion can be used to convert electron configurations into (7x4) images by adding 9 empty columns (_7p, _7d, _7f, _6d, _6f, _5f, _3f, _2d, _2f). In addition, all empty cells have been replaced with zero values.

As an example, let us consider the element Iridium (atomic number 77) that has the electron configurations shown above. The (7x4) image of electron configurations of Iridium is shown in Figure 4 and it represents data of the realigned Table 2, b.

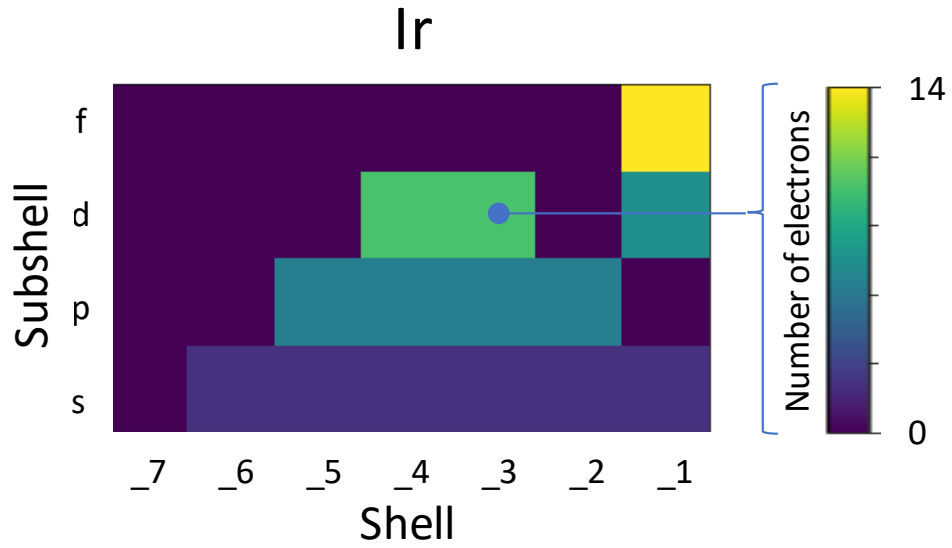

Figure 4. Example of electron configurations presented as (7x4) image for element Iridium

Reshaping input data as (7x4) images allows for experimentation with convolutional neural network algorithms (Dumoulin and Visin, 2018).

## 5 Input Data and Algorithmic Experimentations

Details of the complete design of experiments (DOE) that support research are shown in Table 3.

| Factor | Levels |
|---|---|
| Structure of input dataset | Original / Realigned |
| Number of inputs | 7, 19, 28 (7x4) |
| Normalization of inputs | Total / Individual |
| Linearization | NA / sqrt() |
| Added noise to inputs | NA / Gaussian |
| Machine learning algorithm | Sequential / CNN |

Table 3. Applied design of experiments

It includes the following factors: (1) original and realigned input data; (2) input data that includes: shell data (seven input variables), shell + subshell data (19 input variables), and shell + subshell data reshaped as (7x4) images; (3) total normalization of input variables or individual normalization by each column; (4) with or without input linearization; (5) with or without adding Gaussian noise; (6) applying sequential deep learning or convolutional neural networks algorithms.

The total normalization emphasizes the differences of the maximum number of electrons between shells and subshells. Individual normalization by each column provides for better recognition of outliers, but not a general representation. Rationale of linearization applying square root transformation is based on the maximum numbers of electrons per shell that are in the following quadratic forms: $2=2*1^2$, $8=2*2^2$, $18=2*3^2$, and $32=2*4^2$ (see Table 1, a).

It turns out that the convolutional β-VAE algorithm with realigned (7x4) input images and total normalization with no added noise and without linearization is more accurate compared to the deep sequential algorithms or other combinations of factors. Therefore, all further results are presented applying that setup.

## 6 Application of Convolutional β-VAE for Realigned Electron Configurations

When training autoencoders, it is important to compare how close the decoded outputs are to the inputs. It can be observed that by applying convolutional β-VAE, the decoded images match their inputs quite well despite the above-mentioned challenges of discrete and sparse data (Figure 5).

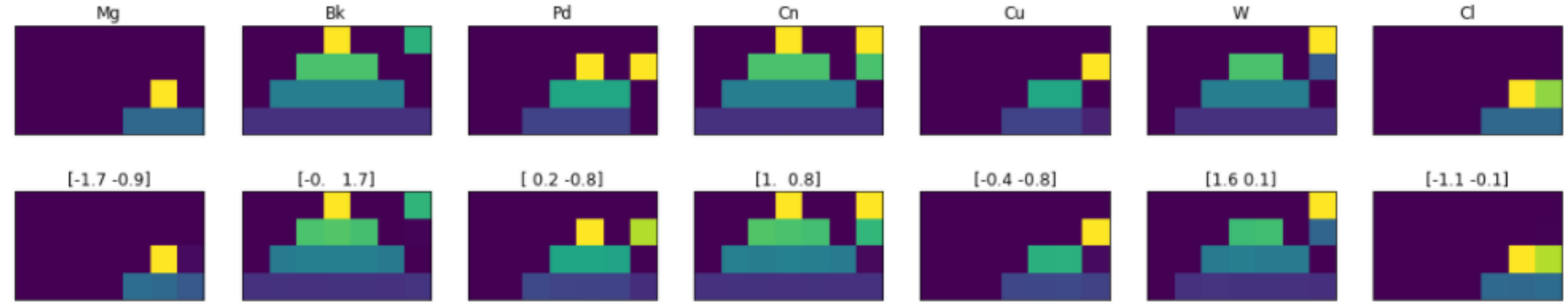

Figure 5. Examples of input images of some elements (top row) and their corresponding decoded images and coordinates on the latent space (bottom row)

Tuning of the beta parameter reveals that small values, such as the selected β=0.03, provide meaningful disentanglement of periods and groups, while increased values destroy disentanglement and, as expected, reduce the accuracy of autoencoding. Completely switching out the Kullback–Leibler regularization term by setting beta to zero, the manifold is severely losing disentanglement power.

## .7 Representation of Elements on the Latent Space Using Convolutional β-VAE

Some characteristics of the chemical elements on the latent space applying convolutional β-VAE are presented in Figure 6. It can be observed that there are clear patterns concerning (a) periods, (b) groups, (c) blocks, and (d) types of elements. For example, blocks "s" and "p" are aligned along radial rays, block "f" on a circular path, and block "d" in between. Elements are ordered by atomic numbers within each period. Mapping of the types of elements is shown in Figure 6, d. It has a pattern that matches the representation of blocks quite well (Figure 6, c). It means that the representation of elements on the encoded latent space recalls the major properties of the periodic table.

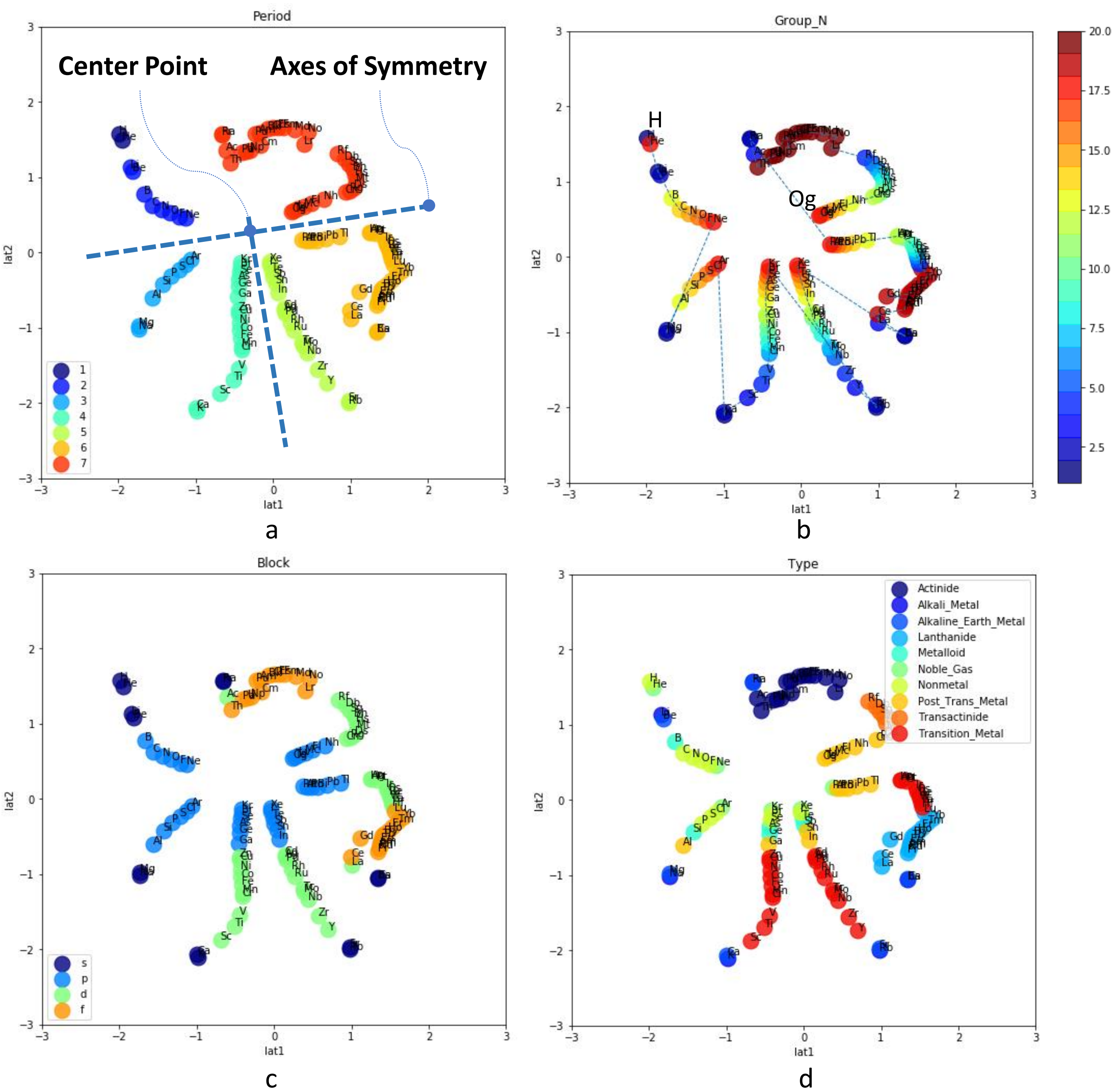

Figure 6. Representation of periods (a), groups (b), blocks (c) and types (d) of elements on the latent space

The unsupervised representation of elements on the latent space shows pairwise symmetries between 2 and 3, 4 and 5, and 6 and 7 periods (Figure 6, a) that allied with equal numbers of elements in these periods (Table 4). The intersection of almost perfectly perpendicular axes of symmetry points to the center of the representation.

Observed symmetries are related to the invariance of quantum numbers of corresponding elements (though there are some exceptions). For example, Zirconium and Titanium are reflections of each other over the axis of symmetry. Both elements have the same quantum number of $^3F_2$. Parity between elements is consistent with atomic number shifts by 8, 18, or 32 (Table 4). Thus, for Zirconium (atomic number 40, period 5) and Titanium (atomic number 22, period 4), the difference between atomic numbers is 18.

Quantum numbers correspond to energy levels of electrons in atoms. Therefore, the observed symmetries mean a chain of discrete relationships:

**Symmetries of chemical elements and periods → Invariance of quantum numbers → Conservation of energy**

This may be a subject for additional study considering Noether's Theorem (Kosmann-Schwarzbach, 2010).

| Period | Number of Elements |
|---|---|
| 1 | $2 = 2x1^2$ |
| 2 | $8 = 2x2^2$ |
| 3 | $8 = 2x2^2$ |
| 4 | $18 = 2x3^2$ |
| 5 | $18 = 2x3^2$ |
| 6 | $32 = 2x4^2$ |
| 7 | $32 = 2x4^2$ |
| | **Total: 118** |

Table 4. Number of elements per period

It should be noted that the radial rays and circular structures of the representation provides rational ground to apply a polar coordinate system that will be discussed later.

In addition to categories, the physical properties of the elements can be mapped on the latent space as well. An example for the melting point temperature is shown in Figure 7, a.

Along representations of properties shown as mappings in Figures 6 and 7, the pattern of the manifold on the latent space can be considered for detecting some irregularities. Clearly, green and red points on the latent space representation shown in Figure 7 (b) are irregularities. It turns out that these points belong to elements of periods 6 and 7, where Madelung's rule is violated (Madelung, 1936; Meek and Allen, 2002). It means that the latent space representations can be used as a research instrument for finding outliers, violations, and irregularities. For continuous cases, research can be supported by unsupervised segmentation of the latent space based on weight-of-evidence of posterior versus standard isotopic 2D normal densities (Glushkovsky, 2020).

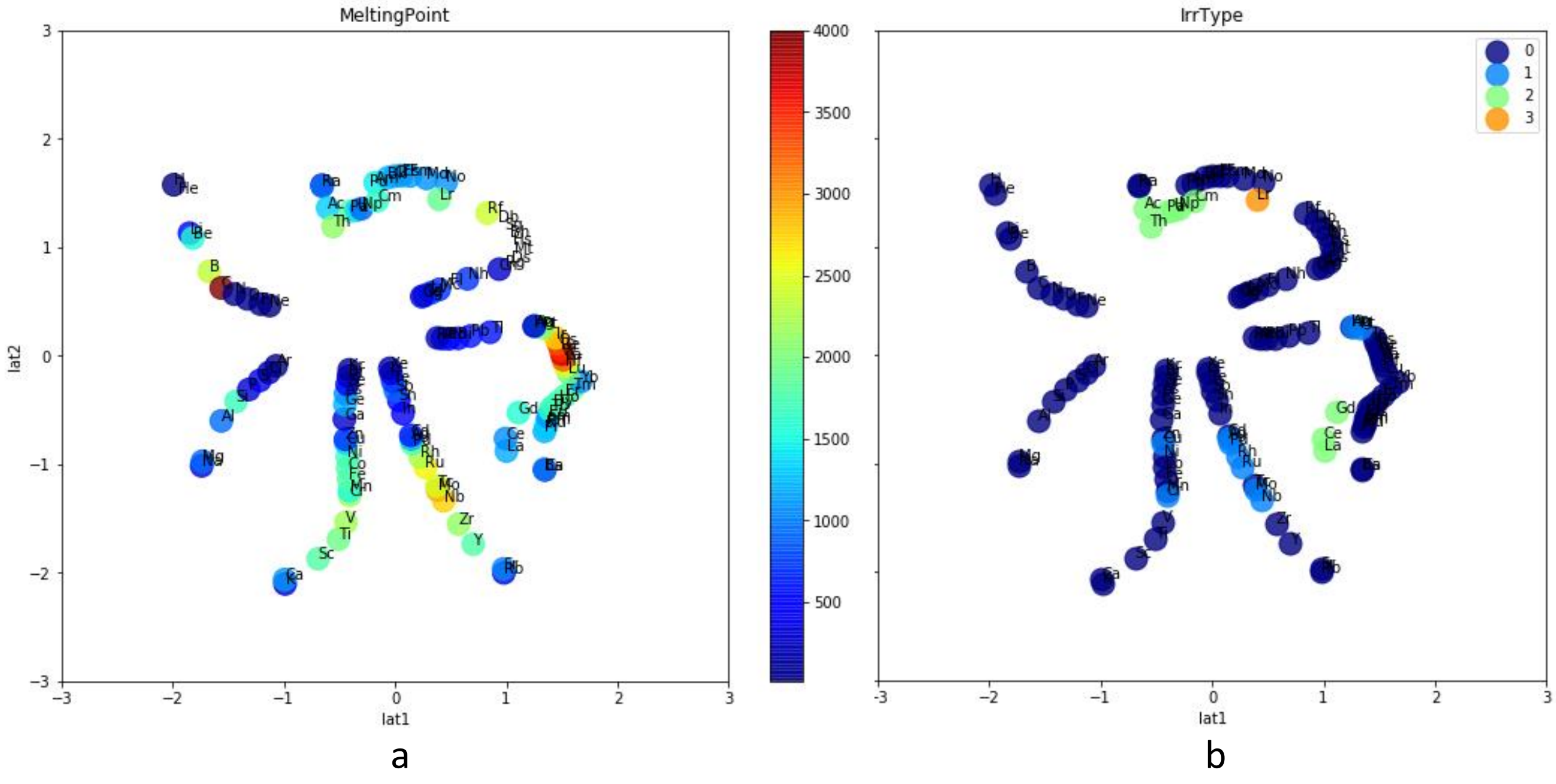

Figure 7. Representation of melting point temperature (a) and violations of Madelung's rule (b) mappings on the latent space

## 8 Representation of Elements on the Polar Coordinate System of the Latent Space

Intersection of two perpendicular axis of symmetry (Figure 6, a) locates the center point of the latent space. Selecting the point of view at that intersection, it can be revealed that periods are separated by an angle, while groups are disentangled along radial rays (Figure 6, a and b). Therefore, it makes sense to apply transformation of the orthogonal coordinate system of the 2D latent space into a polar one. Figure 8 presents such transformation.

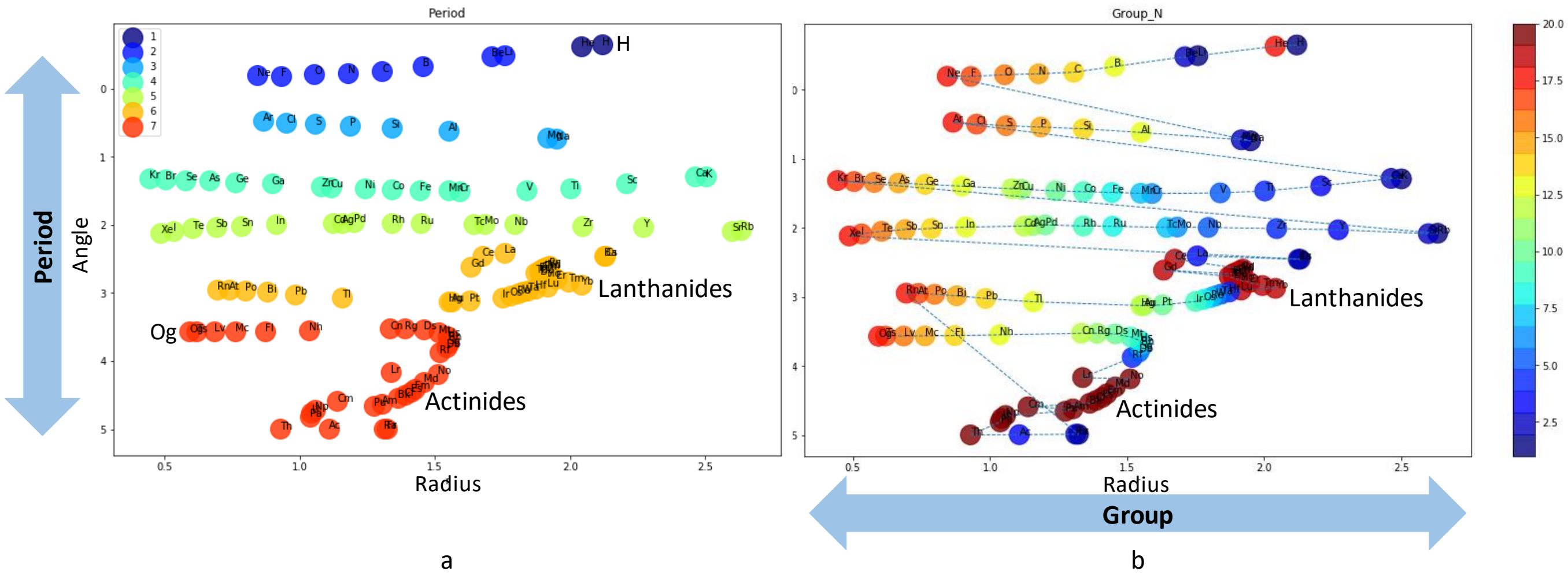

Figure 8. Representation of elements on the polar coordinate system of the latent space with period (a) and group (b) mappings

As expected, it separates periods along angles and disentangles groups along radiuses. Except some distortions among lanthanides and actinides, the polar representation is quite similar to the traditional tabular one (Figure 1, b) but has additional information concerning distances between elements especially for those that belong to the neighboring elements along atomic numbers.

## 9 Generation of Electron Configurations by β-VAE

Generative VAE can produce outputs for any location including empty spaces on the latent space (Kingma and Welling, 2014). Examples of such generation are shown in Figure 9.

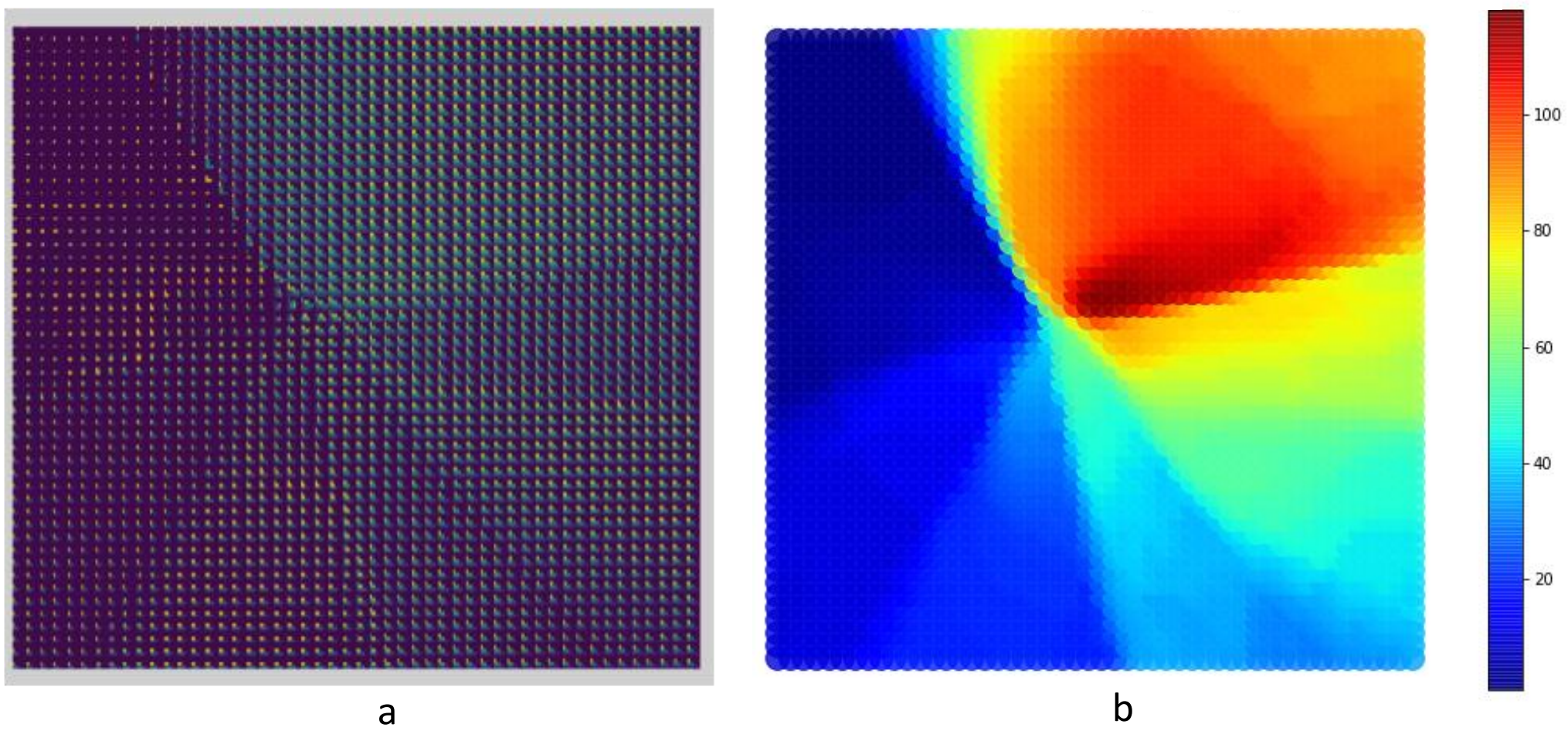

Figure 9. Grid (50x50) of decoded images by convolutional β-VAE on the latent space (a) and their atomic numbers (b)

It can be observed that there are distinct areas presented in Figure 9 (b). The separation between the areas matches the axis of symmetry that has been discussed earlier (Figure 6, a). The central point is located at the intersection between the axes of symmetry where there is the highest gradient of atomic numbers along one axis and a slow change in a perpendicular direction. It can be recognized as a bifurcation point having an atomic number on an average level around 59.

The generation and decoding by β-VAE can be seen as a synthesis of artificial elements producing new electron configurations. However, this generation is not constrained by rules that are present in real physical objects, such as the maximum number of electrons per orbitals or Madelung's order. Furthermore, having a discrete case of input data, that generation is controversial concerning points that are different from the real encoded electron configurations of 118 elements.

Acknowledging these restrictions, the application of the generative properties of the β-VAE may still be helpful as discussed below.

## 10 Discrete Input Data Leads to Supervised Learning

From a data point of view, some discrete cases are unique instances having underlying data characteristics that prevent their continuity. For example, the electron configurations are constrained by some rules as mentioned in the previous paragraph. Acknowledging the existence of such restrictions, it may be possible to learn some insights by observing real input data versus artificial decoded points from the generated grid on the latent space. It means flipping a limitation into a learning opportunity.

The study testing hypothesis that data inputs itself have discriminatory power in classifying real and artificial cases has been performed. It includes a simple supervised machine learning task with automatically identified labels (Figure 10). To improve the accuracy of the study the input values of the real elements have been used instead of the encoded-decoded ones as shown in Figure 10.

If that supervised learning reveals considerable ability to classify real versus artificial cases, then it suggests that the input data includes some inherent patterns that are responsible for the discretization. If not, it means that there are external factors. Similarly, in mathematics, it is worthwhile to discover a proof that the problem has a solution. Of course, to find the solution is even more essential, but knowing that a solution exists is the first step towards finding it.

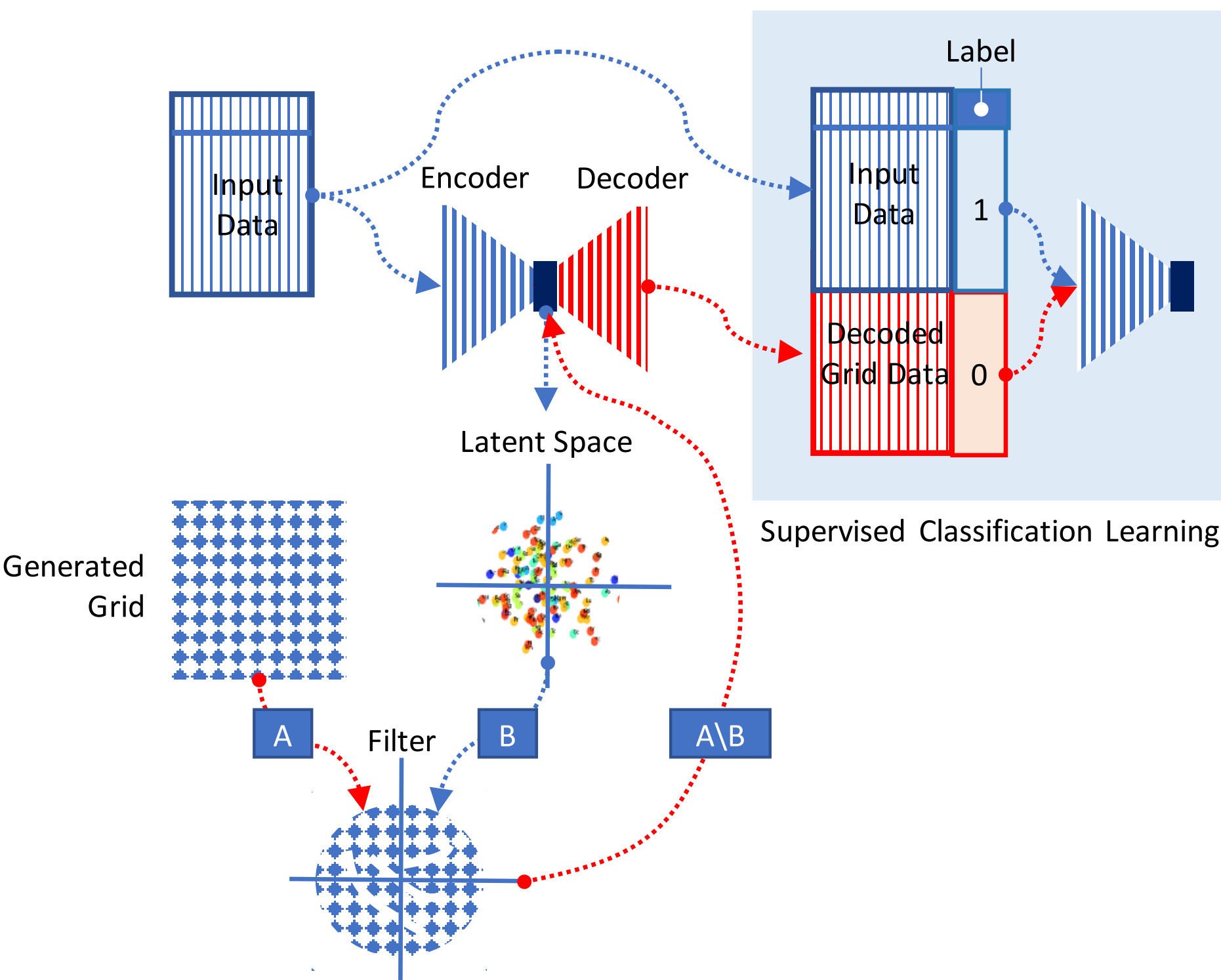

Figure 10. Supervised learning classifying real discrete inputs and generated decoded data

To support such a study, the electron configurations of real chemical elements have been mixed with artificial ones that have been decoded by the trained model from a generated (50x50) grid on the latent space (Figure 9, a). Then this grid has been subjected to the following filters. To start, the real and generated points to be distanced within the range [0.2; 0.7]·σ on the latent space. The lower limit has been set to eliminate cases with poor differentiation and, therefore, to reduce false positive cases, and the higher limit has been set in order to exclude cases that are too far from the trained domain. The finalized manifold of the generated and screened points on the latest space is shown in Figure 11 (blue dots) along with the representation of the real elements (red dots).

Overall, the trained classification model in our example has 81% accuracy based on the test dataset. However, it has only two out of 118 (1.6%) false negative cases misclassifying Mg and Ca elements as artificial.

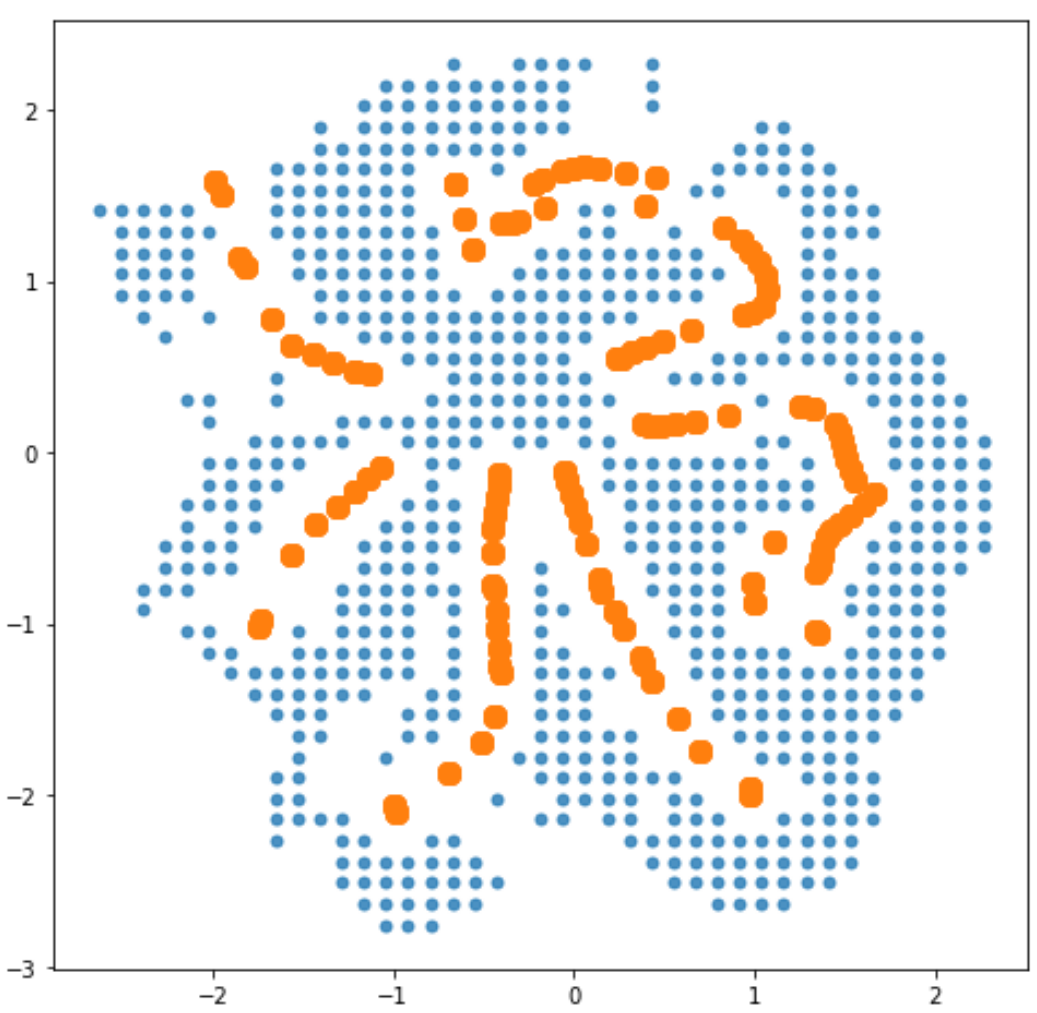

Figure 11. Encoded representations of real elements (red dots) and generated grid after filtering (blue dots) on the latent space

In total, there are 53 false positive cases out of 171 generated points after data deduping (Figure 11, blue dots). These misclassified false positive cases include 26 decoded generations where electron configurations exactly match configurations of the real elements despite the applied filter. After altering these matched cases, the accuracy of classifying artificial cases became 84.2%. For comparison: the accuracy of classifying artificial cases based on Madelung's rule is 94.7%.

Summarizing the obtained results, it turned out that:

- The electron configurations include information that allows for considerable classification of the real and generated cases.
- The trained model has only 10.5% less accuracy than classification based on Madelung's rule of the order of orbital energies.
- The violations of Madelung's rule affect 16.9% real elements. It means that the trained model outcome is comparable to the "accuracy" of the nature.

In the discussed classification study, the applied convolutional neural network model is not easily interpretable and cannot be used for explicit discovery of the generalized rules, such as Madelung's order. Now, knowing that input data of electron configurations include inherent patterns that distinguish real chemical elements from generated ones, let us try to discover the patterns using an unsupervised neural network that encodes input variables.

## 11 Dual Representation by Autoencoders

Conventionally, autoencoders represent observations on the latent space. It means that each observation has a corresponding encoded point on the latent space manifold with a reduced number of description variables down to the bottleneck size (in our examples to 2D vector). Of course, any point on the latent space can be decoded but let us consider only the encoded input set. That representations reveal relationships between input observations, if any.

It has been shown above that unsupervised representation of chemical elements on the latent space based on electron configurations disentangles properties, such as periods, blocks, groups, and types (Figure 6). If autoencoders can successfully represent relationships of observations on the latent space, then, it is logical to assume that autoencoders may be helpful in representing input variables as well. Representation of input variables can be seen as an analogue to variable clustering that addresses collinearity and associations (Lee, 2011).

So, the question arises: "Is it possible to apply autoencoders to reveal relationships between input variables?". It means utilizing autoencoders for dual representation: observations and input variables.

Using simple transposing of input data, it is possible to swap observations and variables. However, the assumed condition of successful training of conventional autoencoders is that the number of observations is significantly larger than the number of input variables and normally enough to split data for training and testing. Also, the number of input variables should be reasonably limited ensuring for convergence of an encoding-decoding learning process. Obviously, these conditions become problematic for the transposed dataset. To trick the former issue, data duplication can be applied. Usefulness of duplication has been shown above addressing autoencoding of discrete electron configurations. The later issue can be mitigated by applying randomized subsampling to bring the number of columns after transposing below practical limits or segmenting the original dataset by defined criteria of interest.

The illustration of dual representation is shown in Figure 12.

The representation of variables can lead to discovery of meaningful patterns along input variables. For example, let us apply the described approach of variables representation (Figure 12, b) for original electron configurations data that includes 19 variables and 118 observations (Table 2, a).

## 12 Discovering Madelung's Rule through Representation of Input Variables on the Latent Space

The experimental data after transposing the original table includes 9,500 (19x500) rows, providing 500x times the duplication and 118 columns. The applied algorithm is sequential deep learning β-VAE described in the Appendix.

The encoded representation of 19 original variables on the latent space is shown in Figure 13.

It can be observed that the representation has a circular path with the gap between the first "1s" and the last "7p" electron orbitals. Posterior probability densities have equal values along the circular path. Considering the clockwise order of variables in Figure 13, it turned out to exactly match Madelung's rule (Madelung, 1936). The latest dictates order of filling the electron orbitals.

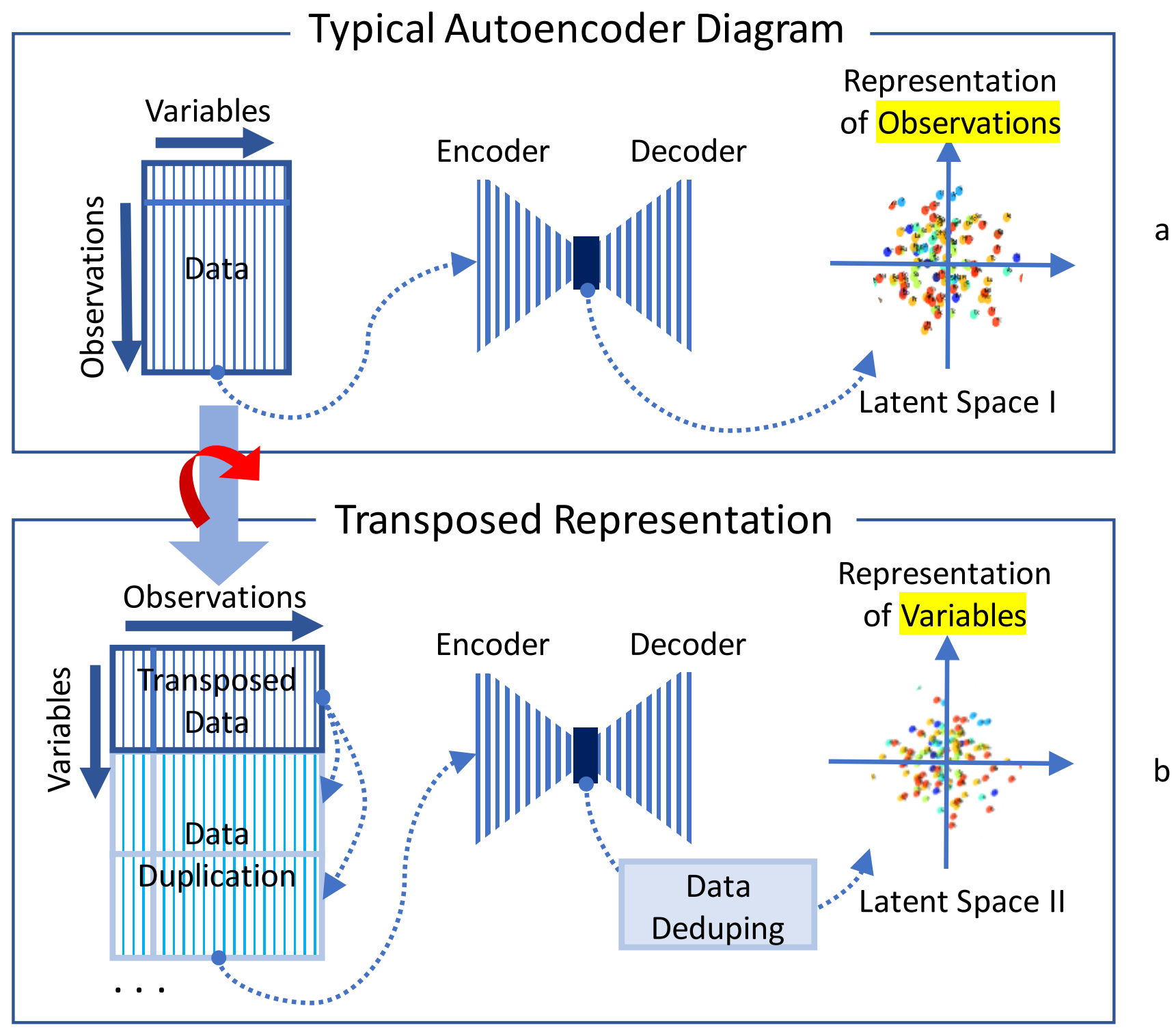

Figure 12. Dual representation on the latent spaces: (a) observations and (b) variables

Furthermore, the latent space representation reflects irregularity of the gap between "5s" and "4d" points that is highlighted in Figure 13. It corresponds to the highest number of Madelung's rule violations in the fifth period for Nb, Mc, Ru, Rh, Pd, and Ag elements, where there is the smallest gap between "5s" and "4d" subshells (Meek and Allen, 2002). Overall, the rule is not perfect and there are some violations affecting about 17% of the elements (see Figure 7, b).

A similar result has been obtained by applying one-dimensional latent representation with β=0.1.

The representation of variables not only rediscovers Madelung's rule but also pointes to the most problematic area concerning that rule violation.

## 13 Alternative Input Features and Representations

Nineteen input features (Table 2) can be reengineered emphasizing different aspects of electron configurations.

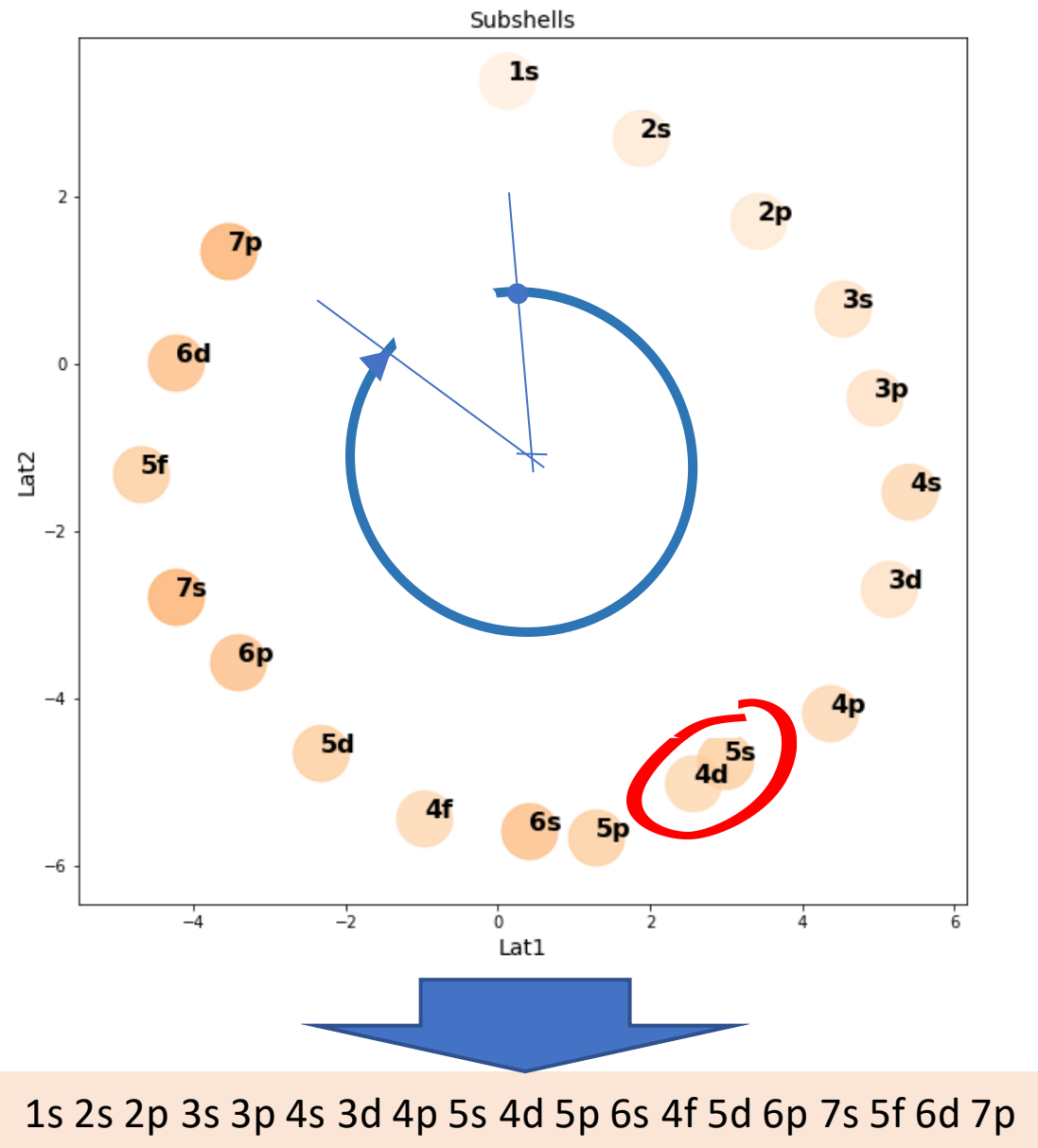

Figure 13. Representation of variables on the latent space, where the clockwise sequence in manifold matches Madelung's rule of filling electron orbitals

The established abbreviation of electron configurations uses the preceding noble gas notation (Masterton and Hurley, 2009). For example, the Iridium element structure is presented as [Xe]$4f^{14}5d^{7}6s^{2}$, where superscripts represent numbers of electrons that exist in the element in addition to the preceding noble gas Xenon. Noble gases have a unique property of containing the maximum possible number of electrons in outermost valence shells. It means that they have reached stable states.

Analogously, it is possible to present electron configurations of elements by comparing them to the following noble gas structures. In this case, notation of Iridium will be presented as [Rn]$5d^{3}6p^{6}$, where figures in superscripts equal the number of vacancies in the valence orbitals of the element compared to the following noble gas Radon. In this alternative abbreviation, the noble gas and the element of interest are from the same period.

The alternative abbreviation allows for reengineering of input features as shown in Table 5, (a) and (b), based on original data that is presented in Table 2, (a) and (b), consequently. The reengineered features have input vectors of zero values for all noble elements.

| | | Shell | | | | | | | | | | | | | | | | | | |
|---|---|---|---|---|---|---|---|---|---|---|---|---|---|---|---|---|---|---|---|---|
| Element | | 1 | 2 | | 3 | | | 4 | | | | 5 | | | | 6 | | | 7 | |
| Num | Name | 1s | 2s | 2p | 3s | 3p | 3d | 4s | 4p | 4d | 4f | 5s | 5p | 5d | 5f | 6s | 6p | 6d | 7s | 7p |
| 1 | H | 1 | | | | | | | | | | | | | | | | | | |
| 2 | He | 0 | | | | | | | | | | | | | | | | | | |
| 3 | Li | 0 | 1 | 6 | | | | | | | | | | | | | | | | |
| 4 | Be | 0 | 0 | 6 | | | | | | | | | | | | | | | | |
| 5 | B | 0 | 0 | 5 | | | | | | | | | | | | | | | | |
| 6 | C | 0 | 0 | 4 | | | | | | | | | | | | | | | | |
| 7 | N | 0 | 0 | 3 | | | | | | | | | | | | | | | | |
| 8 | O | 0 | 0 | 2 | | | | | | | | | | | | | | | | |
| 9 | F | 0 | 0 | 1 | | | | | | | | | | | | | | | | |
| 10 | Ne | 0 | 0 | 0 | | | | | | | | | | | | | | | | |
| 11 | Na | 0 | 0 | 0 | 1 | 6 | | | | | | | | | | | | | | |
| 12 | Mg | 0 | 0 | 0 | 0 | 6 | | | | | | | | | | | | | | |
| 13 | Al | 0 | 0 | 0 | 0 | 5 | | | | | | | | | | | | | | |
| 77 | Ir | 0 | 0 | 0 | 0 | 0 | 0 | 0 | 0 | 0 | 0 | 0 | 0 | 3 | | 0 | 6 | | | |
| 116 | Lv | 0 | 0 | 0 | 0 | 0 | 0 | 0 | 0 | 0 | 0 | 0 | 0 | 0 | 0 | 0 | 0 | 0 | 0 | 2 |
| 117 | Ts | 0 | 0 | 0 | 0 | 0 | 0 | 0 | 0 | 0 | 0 | 0 | 0 | 0 | 0 | 0 | 0 | 0 | 0 | 1 |
| 118 | Og | 0 | 0 | 0 | 0 | 0 | 0 | 0 | 0 | 0 | 0 | 0 | 0 | 0 | 0 | 0 | 0 | 0 | 0 | 0 |

a

| | | Shell | | | | |
|---|---|---|---|---|---|---|
| Element | | _1 | | | | |
| Num | Name | _1s | _1p | _1d | _1f | Period |
| 1 | H | 1 | 0 | 0 | 0 | 1 |
| 2 | He | 0 | 0 | 0 | 0 | 1 |
| 3 | Li | 1 | 6 | 0 | 0 | 2 |
| 4 | Be | 0 | 6 | 0 | 0 | 2 |
| 5 | B | 0 | 5 | 0 | 0 | 2 |
| 6 | C | 0 | 4 | 0 | 0 | 2 |
| 7 | N | 0 | 3 | 0 | 0 | 2 |
| 8 | O | 0 | 2 | 0 | 0 | 2 |
| 9 | F | 0 | 1 | 0 | 0 | 2 |
| 10 | Ne | 0 | 0 | 0 | 0 | 2 |
| 11 | Na | 1 | 6 | 0 | 0 | 3 |
| 12 | Mg | 0 | 6 | 0 | 0 | 3 |
| 13 | Al | 0 | 5 | 0 | 0 | 3 |
| 77 | Ir | 0 | 6 | 3 | 0 | 6 |
| 116 | Lv | 0 | 2 | 0 | 0 | 7 |
| 117 | Ts | 0 | 1 | 0 | 0 | 7 |
| 118 | Og | 0 | 0 | 0 | 0 | 7 |

b

Table 5. Reengineered input features as vacant number of electrons to reach stable states: (a) based on original data (Table 2, a) and (b) based on realigned data towards valent orbitals (Table 2, b)

It should be noted that the reengineered input data in Table 5 (b) has only four remaining informative features (see Appendix, 1). The other fifteen columns have only zero values and have been dropped. Reengineering provides dramatic reduction of the input dimensionality. However, most of the elements that belong to paired periods (Table 4) have identical input vectors, for example, Li and Na (Table 5, b). To restore the uniqueness of the input features of all 118 elements, it is necessary to include additional variable separating periods, such as period or atomic numbers.

Experiments with both additional variables show quite similar results and, therefore, only results with period number variable are presented.

Furthermore, two approaches of input transformations have been conducted concerning the period number: (1) normalization, and (2) one-hot encoding into seven binary variables.

Focusing on the chemical properties of elements, only electron configurations of the valent orbitals can be included as input features. However, keeping only four variables of shell "_1" in Table 2 (b) will be not enough to uniquely distinguish elements. To solve this issue, the period number has been included as a fifth input variable as discussed above.

Considering Table 2 (b), it can be concluded that keeping only the 11 outermost inputs (_4d, _4f _3s, _3p, _3d, _2s, _2p, _1s, _1p, _1d, _1f) out of the 19 variables it will still allow for unique elements identification. Applying only 11 input variables for β-VAE training, the learned representation of elements on the latent space becomes quite similar to that based on all 19 variables (Figures 6 and 7).

Examples of the element representations on the latent space providing alternative input features of different shapes are shown in Figure 14. It should be noted that representations (a) and (c) are based on only five inputs.

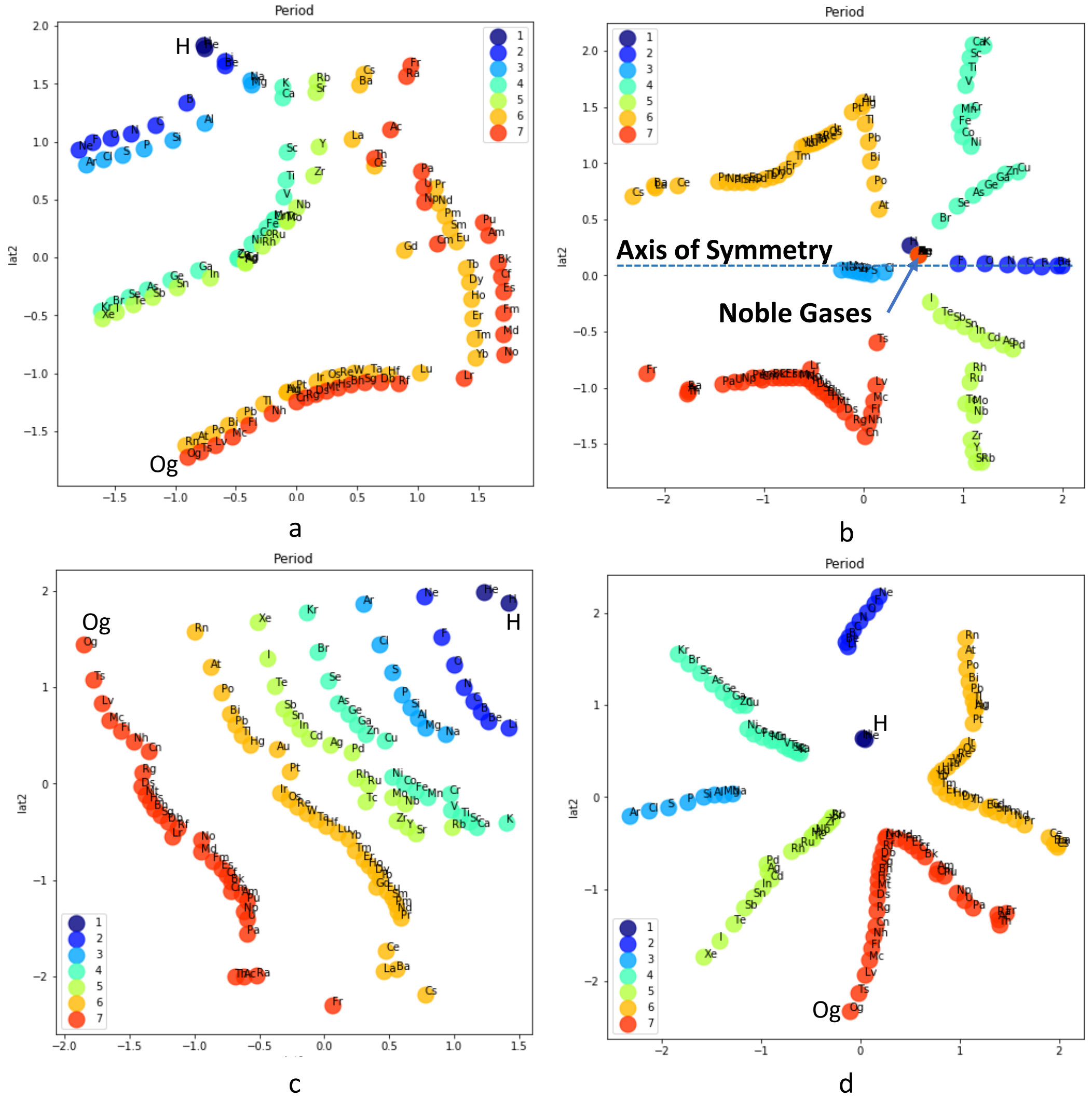

Figure 14. Examples of element representations on the latent space providing different input features:

(a) 5 inputs: valent electron configurations (Table 2, b, shell "_1") and normalized period number
(b) 19 inputs: vacant number of electrons based on original data (Table 5, a)
(c) 5 inputs: vacant number of electrons based on realigned data towards valence orbitals (Table 5, b) and normalized period number
(d) 11 inputs: vacant number of electrons based on realigned data towards valence orbitals (Table 5, b) and one-hot encoded period number

Symmetries between periods and elements, which have been discussed earlier in section 7, can be observed in Figure 14 (b) as reflections over the X-axis (Lat1). On the other hand, corresponding "symmetrical" elements are placed nearby forming three double lines in Figure 14 (a) having identical configurations of valence electrons (though there are some outliers).

It can be observed that input features that emphasise different aspects of electron configurations produce a variety of meaningful patterns on the latent space and can be used for learning, experimentation, or embedding of elements for further research.

## 14 Conclusions and Recommendations for Future Research

The research paper discusses the application of autoencoders as instruments of learning. Representation of chemical elements on the latent space has clear disentanglement of groups, periods, blocks, and types of elements providing discrete and sparse input data of electronic configurations. The output has additional features such as locations and pairwise distances on a 2D coordinate system.

The disentanglement has been triggered after the realignment of electron configurations toward outermost valence orbitals. Electrons of valence orbitals determine the chemical properties of elements. Incorporation of that information provides pivotal impact on the obtained results, without adding new variables or observations. It highlights the importance of feature engineering by simply transforming original data.

Potentially, the realigned electron configurations (Table 2, b, Table 5, a, and Table 5, b) can be helpful in machine learning of materials and not only elements. For example, in the widely used applications of Graph Neural Networks in chemistry where features of atoms are included as node inputs. However, this topic requires further research.

The result has been achieved by considering 118 known elements, i.e., a very limited number of records. It became possible by applying simple duplications of records even without adding Gaussian noise.

Electron configurations presented as (7x4) images, having seven electron shells and four subshells ("s", "p", "d", and "f"), emphasize relationships between input variables. It allows for the application of convolutional β-VAE that improves the accuracy of the trained models.

The unsupervised representation of elements on the latent space reveals pairwise symmetries between 2 and 3, 4 and 5, and 6 and 7 periods that align with the equal number of elements in these periods. The intersection of almost perfectly perpendicular axes of symmetry points to the center of the representation.

Observed symmetries are related to the invariance of quantum numbers of corresponding elements (though there are some exceptions). Parity between elements is consistent with atomic number shifts by 8, 18, or 32 (Table 4).

Quantum numbers correspond to energy levels of electrons in atoms. Therefore, the observed symmetries mean a chain of discrete relationships: Symmetries of chemical elements and periods → Invariance of quantum numbers → Conservation of energy. This may be a subject for additional study considering Noether's Theorem.

The pattern of the manifold on the latent space can be considered in detecting some irregularities. It turned out that outliers of the representation of elements of periods six and seven are cases, where Madelung's rule is violated. It supports the ability of the latent space representations to be used as an instrument for finding outliers, violations, and irregularities.

It has been shown that the representation of elements on the polar coordinate system of the latent space separates periods along angles and disentangles groups by radiuses. Except for some distortions among lanthanides and actinides, the polar representation is quite similar to the traditional tabular periodic table but has additional information concerning distances between elements especially for those that belong to neighboring elements along atomic numbers.

The supervised learning has been set to prove the existence of inherent patterns in electron configurations that differentiate real elements from generated ones that have been decoded by β-VAE from the latent space points. It was performed based on a supervised machine learning task with automatically identified labels. The result of the supervised model training indicates that the electron configurations include information that allows for considerable classification of real and generated cases.

The article addresses the capability of dual representation by autoencoders. Conventionally, autoencoders represent observations of the input data on the latent space. However, by transposing and duplicating original input data, it is possible to represent variables on the latent space as well. The latest can lead to the discovery of meaningful relationships among input variables. Applying that unsupervised learning for transposed electron configuration data, the order of input variables that has been arranged by the encoder on the latent space has turned out to exactly match the sequence of Madelung's rule.

Furthermore, the latent space representation of variables reflects the irregularity of the gap between points "5s" and "4d" that is the most problematic area when it comes to Madelung's rule violations.

## Appendix

## 1 Data Availability

Data for this paper is available at GitHub (https://github.com/alexglush/Element_Electron_Config).

## 2 Architectures of the Models

Architectures of the basic neural network models used in this research have been as follows.

Convolutional β-VAE model (representation of observations):

> Encoder: Flatten input with 28 (7x4) channels; Conv2D (256, kernel_size=(5,2), strides=1, padding='same', activation='relu'); Conv2D (32, kernel_size=(5,2), strides=1, padding='same', activation='relu'); Flatten(896 (7x4x32)); Dense(896 (7x4x32), activation='relu'); two Dense (latent_dim=2).
>
> Decoder: Dense (896 (7x4x32), input_dim=2, activation='relu'), Reshape(7, 4, 32); Conv2DTranspose(32, kernel_size=(5,2), strides=1, padding='same', activation='relu'); Conv2DTranspose(256, kernel_size=(5,2), strides=1, padding='same', activation='relu'); Conv2DTranspose(1, kernel_size=(5,2), padding='same', activation='sigmoid');
>
> Loss=binary_crossentropy+beta*KL; beta=0.03; optimizer='adam'; split between training and testing datasets has been 67%/33%

β-VAE model (representation of observations):

> Encoder: Flatten input with 7 channels; Dense (256, activation='relu'); Dense (32, activation='relu'); two Dense (latent_dim=2).
>
> Decoder: Dense (32, input_dim=2, activation='relu'), Dense (256, activation='relu'), Dense (original_dim=7), activation='sigmoid').
>
> Loss=binary_crossentropy+beta*KL; beta=0.03; optimizer='rmsprop'; split between training and testing datasets has been 67%/33%

β-VAE model (representation of input variables):

> Encoder: Flatten input with 118 channels; Dense (128, activation='relu'); Dense (16, activation='relu'); two Dense (latent_dim=2).
>
> Decoder: Dense (16, input_dim=2, activation='relu'), Dense (128, activation='relu'), Dense (118), activation='sigmoid').
>
> Loss=binary_crossentropy+beta*KL; beta=0.03; optimizer='rmsprop'; split between training and testing datasets has been 67%/33%

Classification model:

Convolutional neural network: Input 28 (7x4); Conv2D (16, kernel_size=(3, 3), strides=(2, 1), padding='same', activation='relu'); Conv2D (8, kernel_size=(3, 3), strides=(2, 1), padding='same', activation='relu'); MaxPool2D (pool_size=(3, 3)); Dropout (0.25); Flatten(8); Dense (original_dim=2), activation='sigmoid')

Loss= categorical_crossentropy; optimizer=Adadelta; metric=accuracy; split between training and testing datasets has been 60%/40%.